\documentclass[letterpaper]{article} 
\usepackage{aaai25}  
\usepackage{times}  
\usepackage{helvet}  
\usepackage{courier}  
\usepackage[hyphens]{url}  
\usepackage{graphicx} 
\usepackage{natbib}  
\usepackage{caption} 
\usepackage{algorithm}
\usepackage{algorithmic}
\usepackage{newfloat}
\usepackage{listings}
\DeclareCaptionStyle{ruled}{labelfont=normalfont,labelsep=colon,strut=off} 
\floatstyle{ruled}
\newfloat{listing}{tb}{lst}{}
\floatname{listing}{Listing}
\title{Foreground-Covering Prototype Generation and Matching\\ for SAM-Aided Few-Shot Segmentation}
\author {
    Suho Park\thanks{Equal contribution}, 
    SuBeen Lee\footnotemark[1], 
    Hyun Seok Seong, 
    Jaejoon Yoo, 
    Jae-Pil Heo\thanks{Corresponding author}
}
\affiliations{
    Sungkyunkwan University \\
    \{shpa06972, leesb7426, gustjrdl95, ujaejoon, jaepilheo\}@gmail.com \\
    \footnotetext{These authors contributed equally to this work}
}

\usepackage{bibentry}
\usepackage[utf8]{inputenc} 
\usepackage[T1]{fontenc}    
\usepackage{url}            
\usepackage{booktabs}       
\usepackage{amsfonts}       
\usepackage{nicefrac}       

\usepackage{array,multirow,graphicx}
\usepackage{float}
\usepackage{amsmath}
\usepackage{mathtools}
\usepackage{hhline}
\usepackage{boldline}
\usepackage{tabularx}
\usepackage{comment}
\usepackage{color}
\usepackage{amssymb}
\usepackage{booktabs}
\usepackage{multicol}

\usepackage{algorithm}
\usepackage{pifont}
\usepackage{soul} 
\usepackage{subcaption}
\usepackage{lipsum}
\usepackage{booktabs}
\usepackage{nicematrix}
\usepackage{kotex}
\usepackage{algorithmic}
\usepackage{color}
\usepackage{multirow}
\usepackage{hyperref}
\usepackage{colortbl}
\begin{document}
\maketitle

\begin{abstract}
We propose Foreground-Covering Prototype Generation and Matching to resolve Few-Shot Segmentation (FSS), which aims to segment target regions in unlabeled query images based on labeled support images. Unlike previous research, which typically estimates target regions in the query using support prototypes and query pixels, we utilize the relationship between support and query prototypes. To achieve this, we utilize two complementary features: SAM Image Encoder features for pixel aggregation and ResNet features for class consistency. Specifically, we construct support and query prototypes with SAM features and distinguish query prototypes of target regions based on ResNet features. For the query prototype construction, we begin by roughly guiding foreground regions within SAM features using the conventional pseudo-mask, then employ iterative cross-attention to aggregate foreground features into learnable tokens. Here, we discover that the cross-attention weights can effectively alternate the conventional pseudo-mask. Therefore, we use the attention-based pseudo-mask to guide ResNet features to focus on the foreground, then infuse the guided ResNet feature into the learnable tokens to generate class-consistent query prototypes. The generation of the support prototype is conducted symmetrically to that of the query one, with the pseudo-mask replaced by the ground-truth mask. Finally, we compare these query prototypes with support ones to generate prompts, which subsequently produce object masks through the SAM Mask Decoder. 
Our state-of-the-art performances on various datasets validate the effectiveness of the proposed method for FSS.
Our official code is available at \href{https://github.com/SuhoPark0706/FCP}{https://github.com/SuhoPark0706/FCP}
\end{abstract}

\section{Introduction}
\label{sec_introduction}
\begin{figure}[t!]
    {\includegraphics[width=0.47\textwidth]{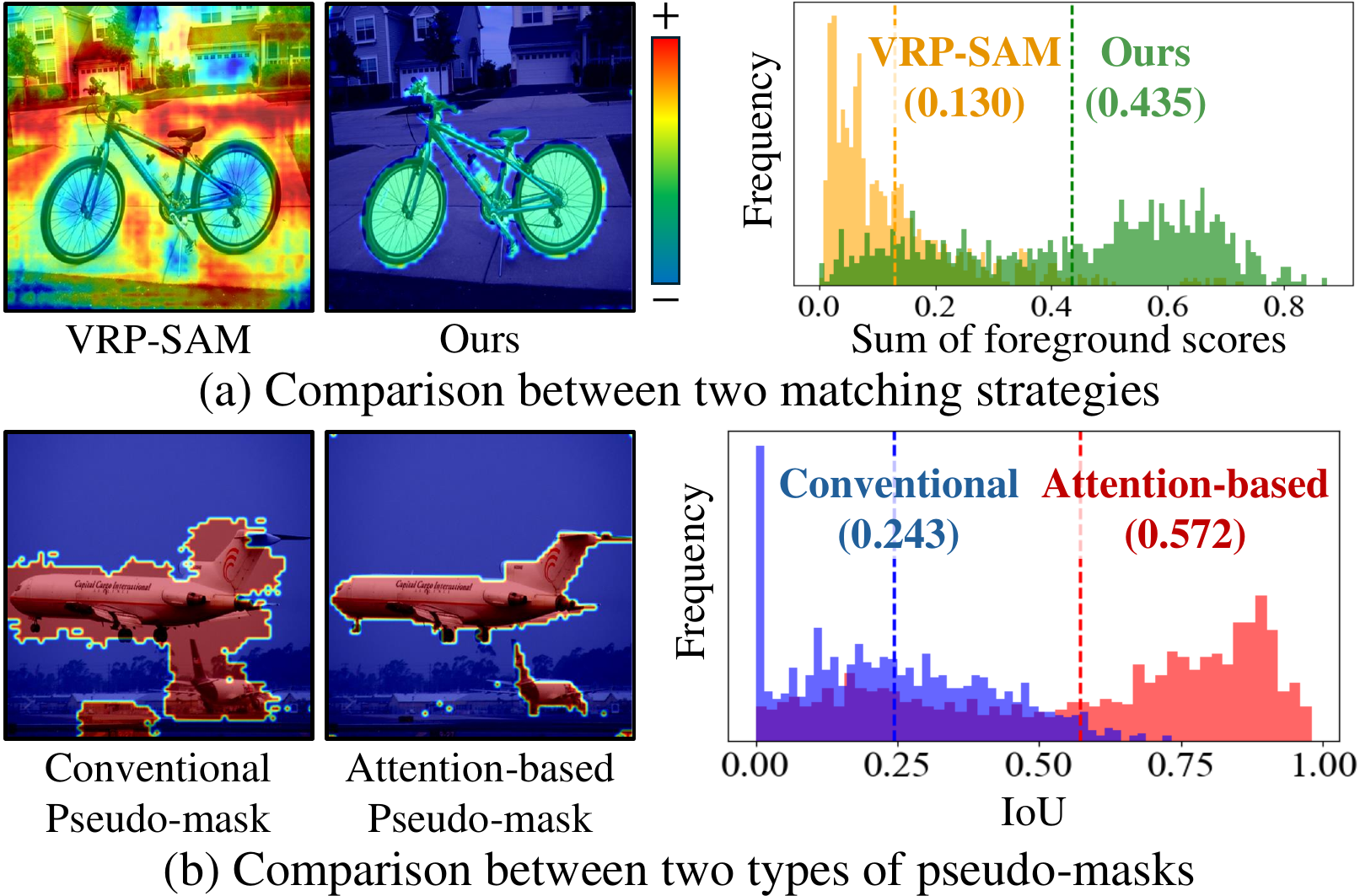}}
    \centering
    \caption{
    Comparison between VRP-SAM and Ours. \textbf{(a)}~(left) We visualize pixel-wise attention maps of query image compared to support prototype. (right) Summing the scores corresponding to the foreground, our prototype-to-prototype matching achieves a higher average score than prototype-to-pixel matching~(VRP-SAM). \textbf{(b)} We compare the conventional and attention-based pseudo masks generated by VRP-SAM and our method. The visualizations and IoU distribution with query foreground validate the effectiveness of our attention-based pseudo-mask.
    }
\label{fig:fig_1}
\end{figure}
Semantic segmentation has achieved remarkable advancements in recent years~\cite{deeplab, segnet, segmenter, maskformer, segformer}.
However, these successes heavily rely on large amounts of labeled data, which require significant human effort. 
To reduce this burden, Few-Shot Segmentation (FSS) has been introduced~\cite{shaban2017one, DCAMA, CyCTR, HDMNet, abcnet}.
FSS aims to segment an unlabeled image, known as the query image, using a small number of labeled images, referred to as the support image.

The Segment Anything Model~(SAM)~\cite{SAM} has demonstrated remarkable versatility across various segmentation tasks.
Although SAM excels at generating object masks with appropriate prompts within an image, effectively utilizing SAM for FSS remains challenging.
This limitation arises from the absence of class consistency across images in SAM, which means it cannot classify the foreground regions of the query image based on the support images.

Recently, VRP-SAM~\cite{VRP-SAM} introduced an alternative approach to leverage SAM for FSS by generating visual reference prompts that encode query foreground pixels corresponding to support foreground pixels, and producing an object mask via the SAM Mask Decoder. 
Specifically, VRP-SAM encapsulates support foreground pixels into learnable tokens, which serve as support prototypes. 
It then generates the visual reference prompts by identifying query pixels that match these support prototypes.
Note that, this process is built upon ResNet feature, ensuring class consistency in distinguishing foreground query pixels based on the support prototypes.

\begin{figure}[t!]
    {\includegraphics[width=0.48\textwidth]{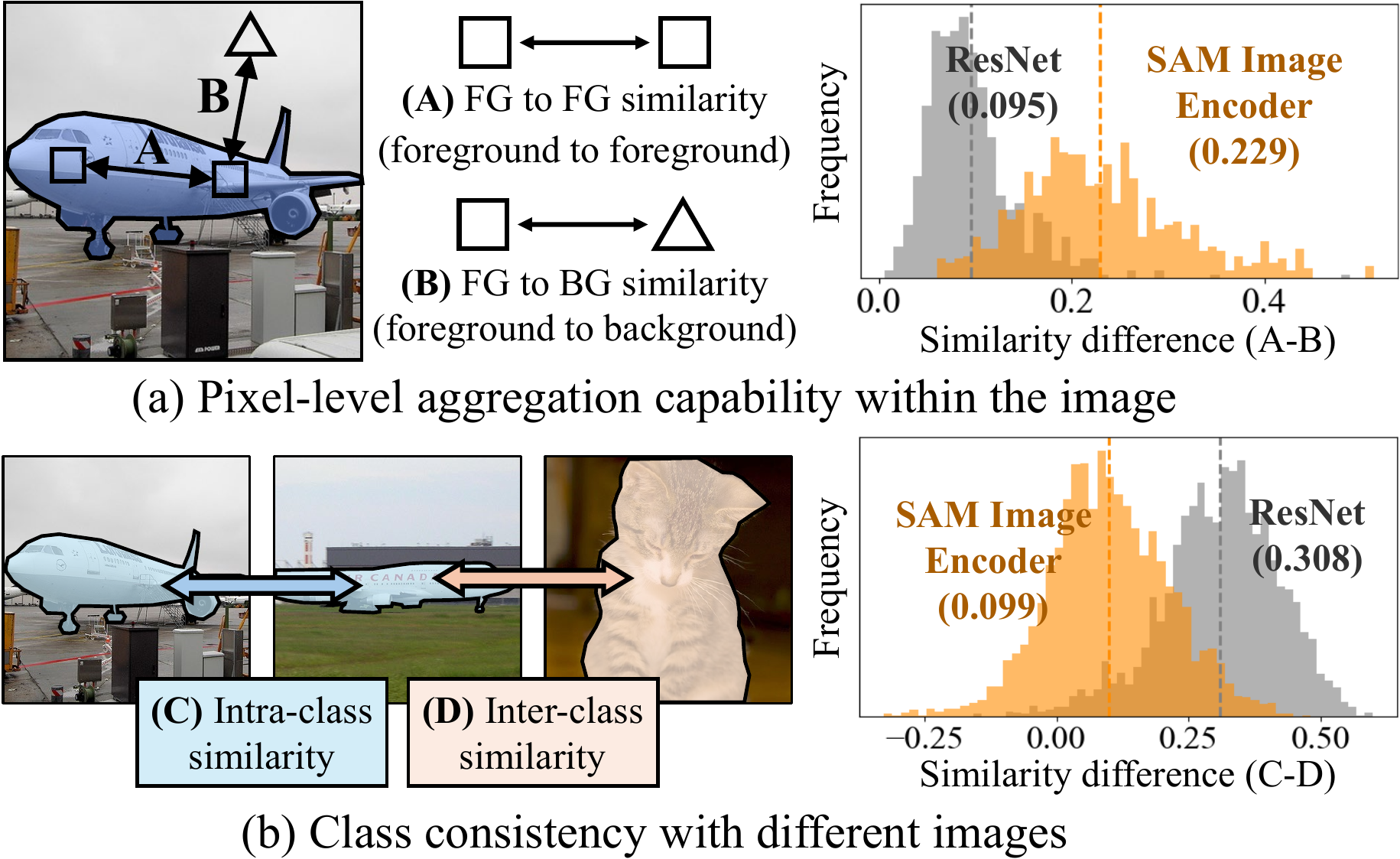}}
    \centering
    \caption{
    Comparison between ResNet and SAM Image Encoder features on 1000 PASCAL VOC images in test dataset, demonstrating their complementary strengths.
    \textbf{(a)} Similarity difference between FG-to-FG and FG-to-BG within the image.
    The higher similarity difference shown in SAM Image Encoder features reflects its superior pixel-level aggregation for prototype construction.
    \textbf{(b)} Difference between the FG-to-FG similarity from the same class (intra-class) and different classes (inter-class).
    Compared to SAM features, ResNet features convey better class consistency across different images.
    }
\label{fig:fig_2}
\end{figure}
Although the pipeline by VRP-SAM is practical, it still has limitations.
Since the relationship between prototypes and pixels is sub-optimal as depicted in Fig.~\ref{fig:fig_1}~(a), this could result in visual reference prompts that either lack sufficient foreground information or contain background elements.
Furthermore, as VRP-SAM uses conventional pseudo-masks based on a simple pixel-pixel similarity, the quality of a conventional pseudo-mask to enhance the query foreground-specific information is low, as described in Fig.~\ref{fig:fig_1}~(b).
Therefore, VRP-SAM struggles to selectively enhance only the query foreground pixels, potentially blurring the distinction between foreground and background pixels. 

To address these issues, we propose Foreground-Covering Prototype Generation and Matching, a method that constructs prototypes from both query and support images that are representative features of each and compares between prototypes to generate reliable visual reference prompts.
Since our method can produce foreground-covering and class-consistent prototypes even within the query, the introduced prototype-to-prototype matching produces more reliable results than prototype-pixel matching as depicted in Fig.~\ref{fig:fig_1}~(a).
This leads to an improvement in the quality of visual reference prompts.
To make prototypes, our approach leverages two types of complementary features: SAM Image Encoder features to group foreground features with its superior aggregation capability~(Fig.~\ref{fig:fig_2}~(a)), and ResNet features, which ensures class consistency~(Fig.~\ref{fig:fig_2}~(b)). 
Specifically, we employ cross-attention with learnable tokens to sequentially encode aggregated foreground information from the SAM features and class-consistent properties from ResNet features.
To effectively aggregate foreground information in the query image, we first roughly guide the foreground features from the SAM features using conventional pseudo-masks. Then, our iterative cross-attention between the SAM features and the learnable tokens ensures a gradual concentration of query foreground features.
For instance, even if the initial pseudo-mask captures only parts of the foreground, the robust aggregation of the SAM features can progressively uncover more of the foreground. Conversely, if the pseudo-mask over-predicts and includes background elements, the SAM features can gradually refine it to better focus more on the foreground.
In the meantime, we discover that the cross-attention weights between learnable tokens and SAM features effectively alternate to the conventional pseudo-mask, as visualized in Fig.~\ref{fig:fig_1}~(b).
Therefore, we utilize this attention-based pseudo-mask to highlight the foreground information in the ResNet features.
Specifically, we infuse foreground-centric class consistency into learnable tokens guided by the attention-based pseudo-mask and ResNet features, to create query prototypes.
This approach allows the prototypes to maintain the class consistency of ResNet features while leveraging the aggregation power of the SAM features.

The process of generating support prototypes is symmetrical to the query prototype generation process, except that the pseudo-masks are replaced by a ground-truth mask.
Consequently, we construct visual reference prompts by matching the query prototypes with the support prototypes.

Overall, our contributions are summarized as follows:
\begin{itemize}
\item{
    We propose a Foreground-Covering Prototype Generation and Matching for few-shot segmentation, which constructs prototypes in both query and support images and compares them to generate visual reference prompts and produce an object mask of query image via the SAM Mask Decoder. 
}
\item{
    To effectively generate foreground-centric prototype, we leverage two types of complementary features: SAM Image Encoder features for its superior aggregation capability and ResNet features having class consistency.
}
\item{
    We propose the attention-based pseudo-mask that can effectively replace the conventional pseudo-mask by leveraging SAM Image Encoder features.
}
\item{
    The effectiveness of our Prototype2Prototype Matching is validated by achieving new state-of-the-art performances across diverse datasets for Few-Shot Segmentation.
}
\end{itemize}

\section{Related Work}
\label{sec_related_work}
\subsection{Vision Foundation Model}
Vision foundation models have demonstrated remarkable adaptability across a wide range of tasks~\cite{CLIP, MaskCLIP, DINO, DenseCLIP}. Among them, Segment Anything Model~(SAM)~\cite{SAM} stands out as a foundation model for segmentation. Due to its promptable design and powerful zero-shot capability, SAM has gained significant attention for its versatility across diverse applications~\cite{Semantic-SAM, SAM-adapter, InpaintAny, SAMmedical} and domains~\cite{SAMmd1, SAM-can, SAMrobust}. 

\subsection{Few-Shot Segmentation}
Few-Shot Segmentation~(FSS) aims to segment the foreground mask for novel classes within a query image with only a few labeled support images.
Specifically, given support images and masks for a specific object, FSS aims to find the mask for the same semantic object within a query image.
FSS has been studied through two mainstreams: prototype-based methods and affinity learning methods.
The prototype-based methods~\cite{ASGNet, Sgone, Panet, ppnet} represent the support foreground as prototypes and utilize them for prediction.
On the other hand, the affinity learning methods~\cite{TBS,sccan, VAT, ea, HSNet, mambafewshot} predict query mask by utilizing pixel-level dense correlation between the support and query images.

Recently, leveraging the zero-shot segmentation capability of the Segment Anything Model (SAM), VRP-SAM~\cite{VRP-SAM} introduced a method to generate appropriate prompts as input for the SAM's mask decoder in the context of FSS.
Nevertheless, the pixel-wise comparison in VRP-SAM has limitations in effectively distinguishing foreground from background. 
In this regard, we show that prototype-wise comparison is more effective in generating foreground-specific prompts, given SAM's strong grouping capabilities within a single image.
\begin{figure*}[t]
    \centering
    {\includegraphics[width=0.98\textwidth]{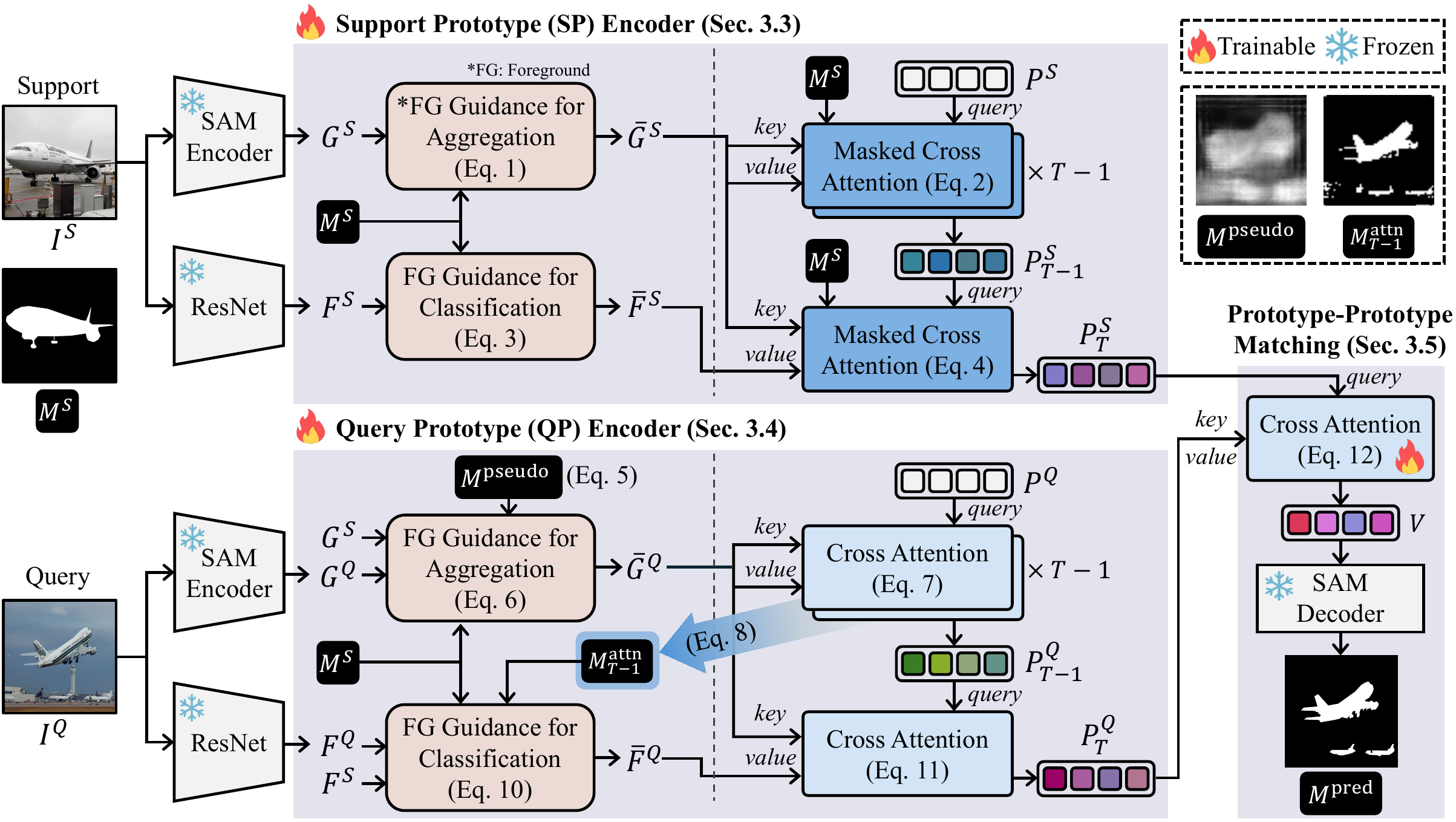} }
    \caption{ 
    Overall procedure of Foreground-Covering Prototype Generation and Matching.
    Given the SAM Image Encoder features $G$, we start by guiding the foreground features using the ground-truth mask for the support $M^S$ and a conventional pseudo-mask for the query $M^\text{pseudo}$, then gather these guided features $\bar{G}$ into learnable tokens $P$ through iterative cross-attention.
    However, SAM features lack class consistency across different images, making it challenging to directly construct prototypes. 
    To address this, we utilize ResNet features $F$ to infuse class-consistent properties into the tokens. 
    We first guide the ResNet features to enhance the foreground-specific information with the ground-truth mask of the support and an attention-based pseudo-mask for the query $M^{\text{attn}}_{T-1}$.
    The attention-based pseudo-mask, benefiting from SAM's high aggregation capability, provides better precision compared to the conventional pseudo-mask, as shown in the upper right. 
    By infusing the class consistency of the guided ResNet features $\bar{F}$ into the learnable prompts, we obtain both support and query prototypes ($P^S_T$ and $P^Q_T$). 
    As a result, the visual reference prompts are generated by matching the query prototypes with the support ones and these prompts are passed to the SAM Decoder to predict a query mask $M^{\text{pred}}$.
    }
\label{fig:fig_3}
\end{figure*}

\section{Method}
\label{sec_method}

\subsection{Problem Definition}
\label{sec_problem definition}
Few-Shot Segmentation (FSS) aims to segment target regions in an unlabeled image by referencing labeled areas in example images.
To accomplish this, FSS typically employs a meta-learning approach. 
In this framework, two separate datasets are used: $\mathcal{D}_{\text{train}}$ for training and $\mathcal{D}_{\text{test}}$ for evaluation. 
These datasets are composed of distinct classes, $\mathcal{C}_{\text{base}}$ and $\mathcal{C}_{\text{novel}}$, which do not overlap~($\mathcal{C}_{\text{base}} \cap \mathcal{C}_{\text{novel}}=\emptyset$). 
Under the above configuration, FSS carries out the training and evaluation through multiple episodes.
In each episode, there are a support set with $K$ labeled images $S=\left\{\left(I^S_i, M^S_i\right)\right\}^{K}_{i=1}$ and a query set with an unlabeled image $Q=\left(I^Q, M^Q\right)$.
Images in both sets contain a randomly sampled category corresponding to the current phase.
$I$ represents the image, and $M$ denotes the ground-truth mask, where 1 indicates the foreground region (the sampled category) and 0 indicates the background region.
As a result, the FSS model is learned to predict the mask of the query set $M^\text{pred}$ based on the support set and query image. 
For simplicity, we assume that the support set contains only one image in the below sections.

\subsection{Overview}

The overall pipeline of our method is described in Fig.~\ref{fig:fig_3}.
We aim to construct both support and query prototypes and match those prototypes to generate reliable Visual Reference Prompts~(VRPs) for SAM Mask Decoder.
To achieve this, we utilize two types of complementary features: SAM Image Encoder features to gather foreground features with its superior aggregation capability~(Fig.~\ref{fig:fig_2}~(a)) and ResNet features to ensure class consistency~(Fig.~\ref{fig:fig_2}~(b)). 
In the following, we will describe each component of our method: Support Prototypes, Query Prototypes, and Prototype-Protype Matching.

\subsection{Support Prototypes}
To produce reliable VRPs by matching support and query prototypes, we first generate Support Prototypes~(SPs) that encode foreground information from the support features.
Since the SPs will be utilized to distinguish foreground and background information of the query features, they should focus on object-specific details and maintain class consistency with other images.
To achieve this, we aggregate foreground information with the SAM features, and then infuse the class consistency of the ResNet features into those aggregated ones.

Given support SAM features $G^S \in \mathbb{R}^{C\times H \times W}$, we first guide foreground features using the ground-truth mask for support $M^S \in \mathbb{R}^{H \times W}$ to facilitate foreground aggregation, as follows:
\begin{align}
    \label{eq:support_SAM_augment}
    \bar{G}^S &=\text{Conv}^G(\text{Concat}(G^S, M^S, \text{MP}(G^S, M^S) )),
\end{align}
where $\bar{G}^S\in\mathbb{R}^{C\times H \times W}$ is guided SAM features while $\text{Conv}(\cdot)$, $\text{Concat}(\cdot)$, and $\text{MP}(\cdot)$ represent a $1\times 1$ convolution layer, concatenate function, and mask-average pooling with expansion, respectively. 
Then, we gather foreground information of those guided features into $N$ learnable tokens $P^S \in \mathbb{R}^{N\times C}$ by employing a cross-attention methodology.
This aggregation process leverages the high pixel-level class consistency within a single image of SAM features and is implemented in a progressive refinement strategy over $T-1$ steps.
Step by step, we encode foreground-specific information $T-1$ times by repeatedly applying cross-attention between the learnable tokens and the SAM features. 
Formally, the $t$-th aggregated support SAM features with the learnable tokens $P^S_t \in \mathbb{R}^{N \times C}$ is defined as follows:
\begin{equation}
    \begin{split}
        \label{eq:SPP_SAM}
        P^S_t = \text{MaskedCrossAttn}(P^{S}_{t-1}, \bar{G}^S, \bar{G}^S ; M^S),
    \end{split}
\end{equation}
where $\text{MaskedCrossAttn}(\cdot)$ denotes cross-attention with a mask condition while $P^{S}_{0}$ is the same with the initial learnable tokens $P^S$.

However, these aggregated features struggle to distinguish foreground and background information of the query features since SAM features lack class consistency with different images compared to ResNet features~(Fig.~\ref{fig:fig_2}).
Therefore, we adopt ResNet features for support $F^S \in \mathbb{R}^{C \times H \times W }$ to grant class-consistent properties to aggregated support features.
To facilitate this, we guide the support ResNet features with the support ground-truth mask, as follows:
\begin{align}
    \label{eq:support_ResNet_augment}
    \bar{F}^S &=\text{Conv}^G(\text{Concat}(F^S, M^S, \text{MP}(F^S, M^S) )),
\end{align}
where $\bar{F}^S\in\mathbb{R}^{C\times H \times W}$ denotes the guided support ResNet features.
Subsequently, we generate SPs $P^S_T \in \mathbb{R}^{N \times C}$ by applying the cross-attention with the aggregated support SAM features $P^S_{T-1}$, the guided support SAM features $\bar{G}^S$, and the guided support ResNet features $\bar{F}^S$ as \textit{query}, \textit{key}, and \textit{value}, respectively.
This can be formulated as follows:
\begin{equation}
    \begin{split}
        \label{eq:SPP_ResNet}
        P^{S}_T = \text{MaskedCrossAttn}(P^{S}_{T-1}, \bar{G}^S, \bar{F}^S; M^S).
    \end{split}
\end{equation}
Through the above process, we produce SPs that consist of foreground features and possess class consistency with other images.

\subsection{Query Prototypes}
The relationship between the SPs and query ResNet features is still insufficient for a reliable comparison, as described in Fig~\ref{fig:fig_1}~(a).
Therefore, we also construct Query Prototyes~(QPs) that each encode query features in the same category to improve trustworthiness and generate reliable VRPs through prototype-prototype comparison.
Note that, the largest difference between SPs and QPs is the absence of the ground truth mask.

Given query ResNet features $F^Q \in \mathbb{R}^{C \times H \times W}$ and query SAM features $G^Q \in \mathbb{R}^{C \times H \times W}$, the QPs generation process consists of foreground features aggregation with SAM features and infusing class consistency using ResNet features, similar to SPs.
However, unlike the support ground-truth mask, we cannot access the query ground-truth mask.
Therefore, we first compute a conventional pseudo-mask for the query $M^{\text{pseudo}} \in \mathbb{R}^{H \times W}$ by utilizing ResNet features, which have class-consistency between different images, following the VRP-SAM~\cite{VRP-SAM}, as follows: 
\begin{equation}
    \begin{split}
        \label{eq:pseudo-mask}
        {M}^{\text{pseudo}}_{h,w} = \max_{1 \leq h' \leq H, 1 \leq w' \leq W } M^S_{h',w'} ( F^Q_{h,w} \cdot F^S_{h', w'} ).
    \end{split}
\end{equation}
where $(\cdot)$ means the cosine similarity.
Using this pseudo-mask, we guide the query SAM features to assist the foreground aggregation, as follows:
\begin{align}
    \label{eq:query_SAM_augment}
    \bar{G}^Q = \text{Conv}^G(\text{Concat}(G^Q, M^{\text{pseudo}}, \text{MP}(G^S, M^S))),
\end{align}
where $\bar{G}^Q \in \mathbb{R}^{C\times H \times W}$ is the guided query SAM features.
Here, we use $\text{MP}(G^S, M^S)$ instead of $\text{MP}(G^Q, M^{\text{pseudo}})$ since the quality of the pseudo-mask is low.
Subsequently, we employ cross-attention to encode foreground-specific information of guided query SAM features into $N$ learnable tokens $P^Q \in \mathbb{R}^{N \times C}$.
Formally, the $t$-th aggregated query SAM features $P^Q_t \in \mathbb{R}^{N \times C}$ is defined as follows:
\begin{align}
    \label{eq:QPPs_SAM}
    P^Q_t = \text{CrossAttn} ( P^Q_{t-1}, \bar{G}^Q, \bar{G}^Q ),
\end{align}
where $\text{CrossAttn}(\cdot)$ refers to the cross-attention while $P^Q_0$ is the same with the initial learnable tokens $P^Q$.

However, unlike SPs that are guided to focus exclusively on foreground regions using the ground truth mask as shown in Eq.~\ref{eq:SPP_SAM} and \ref{eq:SPP_ResNet}, QPs are likely to encode both background and foreground information due to the absence of the ground truth mask. 
To minimize background information, we explicitly encourage QPs to focus on the foreground by introducing a guide loss.
\newline\indent
First, we define an attention-based mask for query $M^{\text{attn}}_t \in \mathbb{R}^{H \times W}$, which is used for defining the guide loss, as follows:
\begin{equation}
    \begin{split}
    \label{eq:attn_mask}
    M^{\text{attn}}_{t,h,w} = \max_{1 \leq n \leq N} A^Q_{t,n,h,w}
    \end{split}
\end{equation}
where $A^Q_{t} \in \mathbb{R}^{N \times H \times W}$ represents attention weights of Eq.~\ref{eq:QPPs_SAM}.
Then, the guide loss $\mathcal{L}_{\text{guide}}$ can be formulated with the attention-based mask, as follows:
\begin{equation}
    \begin{split}
    \label{eq:loss_guide}
    \!\!\!\!\!\! \mathcal{L}_{\text{guide}} \!=\! \frac{1}{T-1}\! \sum^{T-1}_{t=1}\! \mathcal{L}_{\text{BCE}}(M^\text{attn}_t\!, M^Q) + \mathcal{L}_{\text{DL}} (M^\text{attn}_t\!, M^Q) ,
    \end{split}
\end{equation}
where $M^Q \in \mathbb{R}^{H \times W}$ is the ground truth mask of the query, while $\mathcal{L}_\text{BCE}$ and $\mathcal{L}_\text{DL}$ denote the binary cross-entropy loss and the dice loss, respectively. 

Furthermore, we observe that the attention-based mask during aggregation stages can effectively replace the conventional pseudo-mask.
Thanks to the strong aggregation capability of SAM features, even if the pseudo-mask only captures part of the foreground object, the attention weights progressively reveal more details about the foreground regions.
Therefore, when guiding the query ResNet features $F^Q$ to enhance the foreground-specific information, $(T-1)$-th attention-based mask $M^\text{attn}_{T-1}$ was used instead of the conventional pseudo-mask, as follows:
\begin{align}
    \label{eq:QPPs_ResNet}
    \bar{F}^{Q} = \text{Conv}^F(\text{Concat}(F^Q, M^\text{attn}_{T-1}, \text{MP}(F^S, M^S))),
\end{align}
where $\bar{F}^Q \in \mathbb{R}^{C\times H \times W}$ is the guided query ResNet features.
Consequently, we generate QPs $Q^P_T \in \mathbb{R}^{N \times C}$ by applying the cross-attention, as follows:
\begin{align}
    \label{eq:QPPs_for_ResNet}
    P^Q_T = \text{CrossAttn}(P^Q_{T-1}, \bar{G}^Q, \bar{F}^Q).
\end{align}
Although some QPs may encode background information due to the absence of the ground-truth mask in the cross-attention process (Eq.~\ref{eq:SPP_SAM} and \ref{eq:SPP_ResNet}), they still inherit class consistency from ResNet features and benefit from the strong aggregation capability of the SAM features. 
This facilitates the classification of QPs representing foreground, based on SPs.

\subsection{Prototype-Prototype Matching}
\label{sec_proto_proto}
The SAM Mask Decoder requires a reference in the query image to indicate positive or negative areas for segmenting the target region. 
To provide this reference, we explore QPs corresponding to SPs containing foreground information and produce VRPs $V \in \mathbb{R}^{N \times C}$ based on those results, as follows:
\begin{align}
    \label{eq:VRPs}
    V = \text{CrossAttn}(P^S_T, P^Q_T, P^Q_T).
\end{align}
Then, we convey VRPs and query SAM features to SAM Mask Decoder to predict the query mask $M^{\text{pred}} \in \mathbb{R}^{H \times W}$.
To ensure that the predicted query mask closely matches the ground truth mask, we define a prompt loss $\mathcal{L}_{\text{prompt}}$ between the predicted and ground-truth masks, as follows:
\begin{equation}
    \begin{split}
    \label{eq:loss_prompt}
    \mathcal{L}_{\text{prompt}} =\mathcal{L}_{\text{BCE}}(M^{\text{pred}}, M^Q) + \mathcal{L}_{\text{DL}} (M^{\text{pred}}, M^Q).
    \end{split}
\end{equation}
where $\mathcal{L}_\text{BCE}$ and $\mathcal{L}_\text{DL}$ are the binary cross-entropy and dice loss, respectively. 

Additionally, we introduce an orthogonal loss to ensure that the object prompts represent diverse regions by encouraging each SP and QP to encode distinct information. 
The orthogonal loss $\mathcal{L}_{\text{ortho}}$ is computed as follows:
\begin{equation}
    \begin{split}
        \label{eq:loss_orthogonal}
        \!\!\mathcal{L}_{\text{ortho}} = \frac{1}{(T-1)} \sum^{T-1}_{t=1}\sum_{i \neq j}(A^S_{t,i} \cdot A^S_{t,j} ) + (A^Q_{t,i} \cdot A^Q_{t,j} ),
    \end{split}
\end{equation} 
where ${A}^S_{t,n} \in \mathbb{R}^{H\times W}$ and ${A}^Q_{t,n} \in \mathbb{R}^{H\times W}$ represent the cross-attention weights for $N$-th object prompts from Eq.~\ref{eq:SPP_SAM}, Eq.~\ref{eq:SPP_ResNet}, Eq.~\ref{eq:QPPs_SAM} and Eq.~\ref{eq:QPPs_for_ResNet}, respectively, while $(\cdot)$ is the cosine-similarity.
To sum up, the total loss for training is defined as follows:
\begin{equation}
    \mathcal{L}_{\text{total}} = \mathcal{L}_{\text{prompt}} + \lambda_{\text{ortho}}\mathcal{L}_{\text{ortho}} + \lambda_{\text{guide}}\mathcal{L}^Q_{\text{guide}},
\end{equation}
where $\lambda_{\text{ortho}}$ and $\lambda_{\text{guide}}$ are coefficients.

\section{Experiments}
\label{sec_experiments}

\begin{table*}[!t]\normalsize
 \centering
 \setlength{\tabcolsep}{.6mm}{
 {
\begin{tabular}{l|c|ccccc|ccccc|ccccc|ccccc}
\hline
\multirow{3}{*}{\textbf{Method}} & \multirow{3}{*}{\textbf{IE}} & \multicolumn{10}{c|}{\textbf{PASCAL-5$^i$}} & \multicolumn{10}{c}{\textbf{COCO-20$^i$}} \\ \cline{3-22} 
 & & \multicolumn{5}{c|}{\textbf{1-shot}} & \multicolumn{5}{c|}{\textbf{5-shot}} & \multicolumn{5}{c|}{\textbf{1-shot}} & \multicolumn{5}{c}{\textbf{5-shot}} \\ \cline{3-22}  
 & & F-0 & F-1 & F-2 & \multicolumn{1}{c|}{F-3} & Mean & F-0 & F-1 & F-2 & \multicolumn{1}{c|}{F-3} &  Mean & F-0 & F-1 & F-2 & \multicolumn{1}{c|}{F-3} & Mean & F-0 & F-1 & F-2 & \multicolumn{1}{c|}{F-3} &  Mean \\ \hline
HDMNet
 &\multirow{3}{*}{\rotatebox[origin=c]{90}{VGG}}    & 64.8 & 71.4 &  67.7 &\multicolumn{1}{c|}{ 56.4 } & 65.1 & 68.1 &  73.1 & \textbf{71.8} &\multicolumn{1}{c|}{  64.0 } &  69.3 & 40.7 &  50.6 & 48.2 &\multicolumn{1}{c|}{ 44.0 } & 45.9 & 47.0 & 56.5 & 54.1 &\multicolumn{1}{c|}{ 51.9 } &  52.4
  \\
VRP-SAM$^\dagger$
    & & 69.9 & 73.9 & 67.6 &\multicolumn{1}{c|}{ 62.3 } & 68.4 & 72.8 & 75.0 & 67.5 &\multicolumn{1}{c|}{ 63.2 } & 69.6 & 39.4 & 52.0 & 50.6 &\multicolumn{1}{c|}{ 47.5 } & 47.4 & 43.6 & 57.7 & 54.7 &\multicolumn{1}{c|}{ 51.8 } & 51.9  \\
FCP~(Ours)              &  & \textbf{71.9} & \textbf{75.0} & \textbf{69.8} &\multicolumn{1}{c|}{ \textbf{65.5} } & \textbf{70.5} & \textbf{73.5} & \textbf{76.4} &  71.6 &\multicolumn{1}{c|}{ \textbf{65.8} } & \textbf{71.8} & \textbf{44.5} & \textbf{54.3} & \textbf{53.5} &\multicolumn{1}{c|}{ \textbf{47.7} } & \textbf{50.0} & \textbf{49.0} & \textbf{58.3} & \textbf{55.2} &\multicolumn{1}{c|}{ \textbf{52.3} } & \textbf{53.7} \\ \hline

PFENet
 &\multirow{7}{*}{\rotatebox[origin=c]{90}{ResNet}} & 61.7 & 69.5 & 55.4 &\multicolumn{1}{c|}{ 56.3 } & 60.8 & 63.1 & 70.7 & 55.8 &\multicolumn{1}{c|}{ 57.9 } & 61.9 & 36.5 & 38.6 & 34.5 &\multicolumn{1}{c|}{ 33.8 } & 35.8 & 36.5 & 43.3 & 37.8 &\multicolumn{1}{c|}{ 38.4 } & 39.0 \\
CyCTR         & & 65.7 & 71.0 & 59.5 &\multicolumn{1}{c|}{ 59.7 } & 64.0 & 69.3 & 73.5 & 63.8 &\multicolumn{1}{c|}{ 63.5 } & 67.5 & 38.9 & 43.0 & 39.6 &\multicolumn{1}{c|}{ 39.8 } & 40.3 & 41.1 & 48.9 & 45.2 &\multicolumn{1}{c|}{ 47.0 } & 45.6 \\
SSP             & & 60.5 & 67.8 & 66.4 &\multicolumn{1}{c|}{ 51.0 } & 61.4 & 67.5 & 72.3 & \textbf{75.2} &\multicolumn{1}{c|}{ 62.1 } & 69.3 & 35.5 &  39.6 & 37.9 &\multicolumn{1}{c|}{ 36.7 } & 37.4 & 40.6 & 47.0 & 45.1 & \multicolumn{1}{c|}{ 43.9 } & 44.1 \\
BAM             & & 69.0 & 73.6 & 67.6 &\multicolumn{1}{c|}{ 61.1 } & 67.8 & 70.6 & 75.1 & 70.8 &\multicolumn{1}{c|}{ 67.2 } & 70.9 & 39.4 & 49.9 & 46.2 &\multicolumn{1}{c|}{ 45.2 } & 45.2 & 43.2 & 53.4 & 49.4 &\multicolumn{1}{c|}{ 48.1 } & 48.5 \\
HDMNet       & & 71.0 & 75.4 & 68.9 &\multicolumn{1}{c|}{ 62.1 } & 69.4 & 71.3 & 76.2 & 71.3 &\multicolumn{1}{c|}{ \textbf{68.5} } & 71.8 & 43.8 & 55.3 & 51.6 &\multicolumn{1}{c|}{ 49.4 } & 50.0 & 50.6 & 61.6 & 55.7 &\multicolumn{1}{c|}{ 56.0 } & 56.0 \\
VRP-SAM$^\dagger$     & & 74.5 & 77.3 & 69.5 &\multicolumn{1}{c|}{ 65.8 } & 71.8 & 76.3 & 76.8 & 69.5 &\multicolumn{1}{c|}{ 63.1 } & 71.4 & 44.3 & 54.3 & 52.3 &\multicolumn{1}{c|}{ 50.0 } & 50.2 & 50.5 & 59.5 & 56.9 &\multicolumn{1}{c|}{ 54.9 } & 55.5 \\ 
FCP~(Ours)              &  & \textbf{74.9} & \textbf{77.4} & \textbf{71.8} &\multicolumn{1}{c|}{ \textbf{68.8} } & \textbf{73.2} & \textbf{77.2} & \textbf{78.8} & 72.2 &\multicolumn{1}{c|}{ 67.7 } & \textbf{74.0} & \textbf{46.4} & \textbf{56.4} & \textbf{55.3} &\multicolumn{1}{c|}{ \textbf{51.8} } & \textbf{52.5} & \textbf{52.6} & \textbf{63.3} & \textbf{59.8} &\multicolumn{1}{c|}{ \textbf{56.1} } & \textbf{58.0} \\ \hline
\end{tabular}}}
\caption{Experimental results on the PASCAL-5$^i$ and COCO-20$^i$. VGG and ResNet denote VGG-16 and ResNet-50, respectively. IE means the type of image encoder while $\dagger$ indicates the reproduced version for the fair comparison.}
\label{tab:pascal_table}
\end{table*}

\paragraph{Datasets}  
To validate our work, we utilize PASCAL-5$^i$~\cite{shaban2017one-shot_fss} and COCO-20$^i$\cite{cocofewshot} following the prior works~\cite{VRP-SAM, HSNet, TBS}.
Specifically, PASCAL-5$^i$ consists of 20 classes from PASCAL VOC 2012~\cite{pascal} and SDS~\cite{sds}, while COCO-20$^i$ comprises 80 categories from COCO~\cite{coco} dataset.
Each dataset is divided into 4 folds and each fold does not share categories with other folds.
To evaluate generalization capability, we use one of the folds as novel classes for testing and others as base categories for training.
As a result, we use 15 base and 5 novel classes in PASCAL-5$^i$ while 60 base and 20 novel categories in COCO-20$^i$.
To measure the performance of the model, we randomly sample 1000 support-query pairs from novel classes and compute mean Intersection over Union~(mIoU).

\paragraph{Implementation Details}
Following VRP-SAM, we utilize SAM Image Encoder features and other backbone features with class-consistent properties (VGG and ResNet), both pretrained on ImageNet. 
We employ the AdamW optimizer, adjusted by a cosine annealing schedule. 
For PASCAL-5$^i$, we use 100 epochs and an initial learning rate of 2e-4 for training while 50 epochs and an initial learning rate of 1e-4 in COCO-20$^i$, and the batch size of both datasets is 8. 
The number of learnable tokens is 50 for both support and query. 
Additionally, the number of layers for constructing prototypes is 3, \textit{i.e.}, $T=3$.
The loss coefficients are 0.05 and 0.5 for $\lambda_\text{ortho}$ and $\lambda_\text{guide}$, respectively.

\paragraph{Comparison with State-of-the-Art Methods}
We validate our method using standard benchmarks by comparing it against various FSS models.
These include HDMNet~\cite{HDMNet} and VRP-SAM~\cite{VRP-SAM} with a VGG Image Encoder, as well as PFENet~\cite{PFENet}, CyCTR~\cite{CyCTR}, SSP~\cite{SSP}, BAM~\cite{BAM}, and VRP-SAM~\cite{VRP-SAM} with a ResNet Image Encoder.
As shown in Table.~\ref{tab:pascal_table}, our method achieves new state-of-the-art performance, proving its effectiveness. 
This success is consistent across different datasets, shot numbers, and backbone networks, indicating the robustness of our approach. 
Moreover, the superior performance over VRP-SAM indicates that our generated VRP is more advantageous in producing a higher-quality query mask.

\begin{figure}[t!]
    {\includegraphics[width=0.47\textwidth]{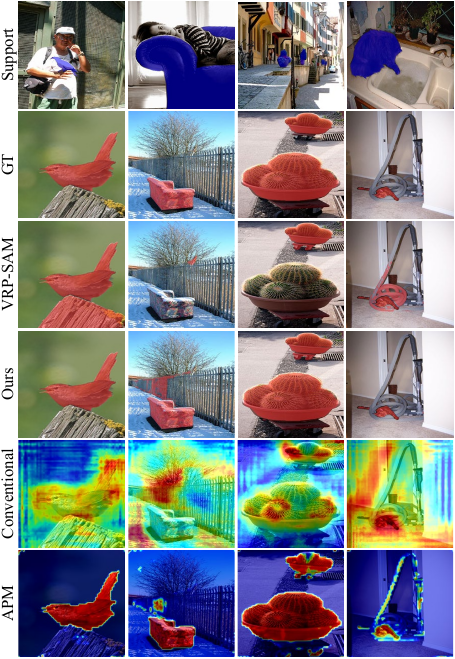}}
    \centering
    \caption{
    Qualitative comparison results of Ours and VRP-SAM on the PASCAL-5$^i$ dataset. Conventional and APM refer to the conventional pseudo-mask of VRP-SAM and our Attention-based Pseudo-Mask, respectively.
    }
\label{fig:fig_4}
\end{figure}

\paragraph{Qualitative Results.}
We compare our qualitative results with VRP-SAM in Figure~\ref{fig:fig_4} and also display both the conventional and attention-based pseudo-masks in VRP-SAM and our method, respectively. 
Our approach consistently outperforms the baseline across different scene types, whether simple (1st and 2nd columns), multi-object (3rd column), or complex (4th column).
This improvement is largely due to the precision of the attention-based pseudo-mask in our method.

\section{Further Analysis}
\label{sec_analysis}

\subsection{Ablation Study}

We perform ablation studies to evaluate the contributions of the main components and key hyperparameters of our method.
Most experiments are progressed with ResNet-50 on PASCAL-5$^i$ with all folds except for $T$ ablation study which conducted with only fold-3 of PASCAL-5$^i$.

\paragraph{Main Components}

Tab.~\ref{tab:ablation_main} presents the results of the ablation study for the main components. 
(a) shows the result of prototype-pixel matching with ResNet-50, which constructs only support prototypes and matches them with the query, which is our baseline model VRP-SAM.
(b) shows the result when using SAM features instead of ResNet features under the same condition.
There is a significant performance drop because SAM features cannot ensure class-consistent properties.
(c) and (d) represent the results when applying our prototype-prototype matching to (a) and (b), respectively. 
For instance, for (c), the inferior pixel-level aggregation of ResNet features causes a performance degradation compared to the baseline.
And for (d), the perfornace still remains poor.
Furthermore, by incorporating prototype-prototype matching by combining SAM's strong aggregation capabilities with ResNet's class-consistent properties in (e), the performance significantly surpasses the baseline.
This emphasizes the importance of using complementary features to construct prototypes.
However, as the conventional pseudo-mask still limits the potential gains, we employ our attention-based pseudo-mask, which leverages SAM's superior aggregation capability.
In specific, (f) further boosts performance, demonstrating that the attention-based pseudo-mask offers precise guidance for the model in locating the target object.

\begin{table}
     \centering
     \begin{tabular}{c | cccc| c| c}
     \hline
      & Res & SAM & PPM & APM & mIoU& $\Delta$\\
     \hline
     (a) & $\checkmark$&            &            &            &71.8&0.0\\
     (b) &             &$\checkmark$&            &            & 66.2 & -5.6 \\
     (c) & $\checkmark$&            &$\checkmark$&            &66.5&-5.3\\
     (d) &             &$\checkmark$&$\checkmark$&            &66.1&-5.7\\
     (e) & $\checkmark$&$\checkmark$&$\checkmark$&            &72.6&+0.8\\       (f) & $\checkmark$&$\checkmark$&$\checkmark$&$\checkmark$&\textbf{73.2}&+\textbf{1.4}\\
     \hline
     \end{tabular}
     \caption{Ablation studies for different components and architectures. Res, SAM, PPM and APM denote ResNet-50, SAM Image Encoder, Prototype-Prototype Matching, and Attention-based Pseudo Mask, respectively.}
     \label{tab:ablation_main}
\end{table}

\begin{figure}[t!]
    {\includegraphics[width=0.47\textwidth]{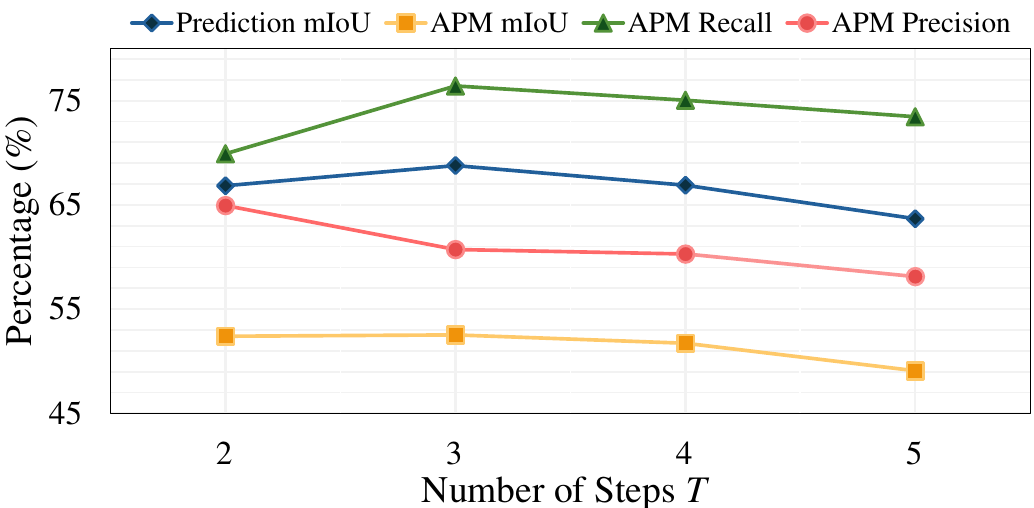}}
    \centering
    \caption{
    Ablation study for varying the number of aggregation steps for prototype construction. Prediction and APM denote our model's query mask prediction and Attention-based Pseudo Mask, respectively.
    }
\label{fig:num_layers}
\end{figure}

\paragraph{Number of Step \textbf{$T$}}
We present the impact of the number of aggregation step for prototype construction in Fig.~\ref{fig:num_layers}.
For instance, 3 steps of prototype construction achieves optimal performance while using more steps leads to a performance drop.
This is because the learnable tokens increasingly focus on smaller areas as they repeatedly apply the softmax function in the cross-attention.
This is further supported by a reduction in the recall of the attention-based pseudo-mask.
A proper $T$ can prevent excessive shrinkage of the pseudo-mask and generate appropriate prototypes.

\begin{table}
     \centering
     \begin{tabular}{ c | c  c  c}
     \hline
      Pseudo-Mask & mIoU & Prec & Rec \\
      \hline
      Conventional &32.4&46.5&53.6\\
      Attention-based & \textbf{60.9} & \textbf{69.1} & \textbf{79.4} \\
     \hline
     \end{tabular}
     \caption{
     Comparison between conventional and attention-based pseudo-masks.
     Prec and Rec denote precision and recall with respect to the ground-truth mask, respectively.
     }
    \label{tab:ablation_pseudo}
\end{table}

\paragraph{Loss Functions}
In addition to the prompt loss $L_\text{prompt}$, which is essential for model training, we employ two supplementary losses: the guide loss $L_\text{guide}$ and the orthogonal loss $L_\text{ortho}$. 
We validate the effectiveness of each loss function, as presented in Tab~\ref{tab:ablation_loss}.
(a) represents the performance when we use only the prompt loss.
(b) and (c) show the results when we adopt the guide loss and the orthogonal loss with the prompt loss, respectively.
In the case of (b), we can improve the performance by guiding the attention-based pseudo-mask to represent the object regions.
For (c), each learnable token can be induced to capture various parts, which brings a slight performance gain.
When we utilize all losses for training as shown in (d), we can get a large performance enhancement.
This verifies the appropriateness of the proposed losses.

\begin{table}
     \centering
     \begin{tabular}{c | ccc | c | c}
     \hline
      & $\mathcal{L}_{\text{prompt}}$ & $\mathcal{L}_{\text{guide}}$ & $\mathcal{L}_{\text{ortho}}$ & mIoU & $\Delta$ \\
      \hline
      (a) & $\checkmark$&            &            &72.3&0.0\\
      (b) & $\checkmark$&$\checkmark$&            &72.7&+0.4\\
      (c) & $\checkmark$&            &$\checkmark$&72.4&+0.1\\
      (d) & $\checkmark$&$\checkmark$&$\checkmark$&\textbf{73.2}&\textbf{+0.9}\\
     \hline
     \end{tabular}
     \caption{Ablation studies about each loss function.}
     \label{tab:ablation_loss}
\end{table}

\subsection{Improvement of Pseudo Mask}

The attention-based pseudo-mask is a key component of our method, used to enhance foreground-specific information. 
As such, its quality directly affects the overall performance of the model.
Table~\ref{tab:ablation_pseudo} compares the quality of pseudo-masks and highlights the performance improvements achieved by our attention-based pseudo mask.
First, the conventional pseudo-mask, computed by a pixel-level cosine similarity, produces a low mIoU with respect to the ground-truth mask.
On the other hand, we introduced our attention-based pseudo-mask by leveraging the strong aggregation capability of the SAM features.
As a result, our attention-based pseudo-mask substantially outperforms the conventional ones, demonstrating the superior foreground-covering capability of our method.

\section{Conclusion}
\label{sec_conclusion}
In this paper, we proposed Foreground-Covering Prototype Generation and Matching, which constructs support and query prototypes and matches them to generate reliable prompts. 
To build prototypes, we utilized two complementary features: SAM Image Encoder for pixel aggregation and ResNet for class consistency.
We guide SAM features using a pseudo-mask for query prototypes, then employ iterative cross-attention to aggregate foreground features into learnable tokens. 
Here, we discovered that the cross-attention weights effectively replace the conventional pseudo mask, allowing attention-based pseudo-masks to guide ResNet features to focus on the foreground.
Then, we encode the guided ones into the learnable tokens to generate class-consistency query prototypes.
Finally, we generated reliable prompts by comparing query and support prototypes, achieving state-of-the-art performance on the few-shot segmentation task.

\appendix

\section*{Acknowledgments}
This work was supported in part by MSIT/IITP (No. 2022-0-00680, 2020-0-01821, 2019-0-00421, RS-2024-00459618, RS-2024- 00360227, RS-2024-00437102, RS-2024-00437633), and MSIT/NRF (No. RS-2024-00357729).

\bibliography{aaai25}

\begin{thebibliography}{41}
\providecommand{\natexlab}[1]{#1}

\bibitem[{Badrinarayanan, Kendall, and Cipolla(2017)}]{segnet}
Badrinarayanan, V.; Kendall, A.; and Cipolla, R. 2017.
\newblock Segnet: A deep convolutional encoder-decoder architecture for image segmentation.
\newblock \emph{IEEE transactions on pattern analysis and machine intelligence}, 39(12): 2481--2495.

\bibitem[{Caron et~al.(2021)Caron, Touvron, Misra, J{\'e}gou, Mairal, Bojanowski, and Joulin}]{DINO}
Caron, M.; Touvron, H.; Misra, I.; J{\'e}gou, H.; Mairal, J.; Bojanowski, P.; and Joulin, A. 2021.
\newblock Emerging properties in self-supervised vision transformers.
\newblock In \emph{Proceedings of the IEEE/CVF international conference on computer vision}, 9650--9660.

\bibitem[{Chen et~al.(2015)Chen, Papandreou, Kokkinos, Murphy, and Yuille}]{deeplab}
Chen, L.-C.; Papandreou, G.; Kokkinos, I.; Murphy, K.; and Yuille, A.~L. 2015.
\newblock Semantic image segmentation with deep convolutional nets and fully connected crfs.
\newblock \emph{ICLR}.

\bibitem[{Chen et~al.(2023)Chen, Zhu, Deng, Cao, Wang, Zhang, Li, Sun, Zang, and Mao}]{SAM-adapter}
Chen, T.; Zhu, L.; Deng, C.; Cao, R.; Wang, Y.; Zhang, S.; Li, Z.; Sun, L.; Zang, Y.; and Mao, P. 2023.
\newblock Sam-adapter: Adapting segment anything in underperformed scenes.
\newblock In \emph{Proceedings of the IEEE/CVF International Conference on Computer Vision}, 3367--3375.

\bibitem[{Cheng et~al.(2022)Cheng, Misra, Schwing, Kirillov, and Girdhar}]{maskformer}
Cheng, B.; Misra, I.; Schwing, A.~G.; Kirillov, A.; and Girdhar, R. 2022.
\newblock Masked-attention mask transformer for universal image segmentation.
\newblock In \emph{Proceedings of the IEEE/CVF conference on computer vision and pattern recognition}, 1290--1299.

\bibitem[{Fan et~al.(2022)Fan, Pei, Tai, and Tang}]{SSP}
Fan, Q.; Pei, W.; Tai, Y.-W.; and Tang, C.-K. 2022.
\newblock Self-support few-shot semantic segmentation.
\newblock In \emph{European Conference on Computer Vision}, 701--719. Springer.

\bibitem[{Hariharan et~al.(2014)Hariharan, Arbel{\'a}ez, Girshick, and Malik}]{sds}
Hariharan, B.; Arbel{\'a}ez, P.; Girshick, R.; and Malik, J. 2014.
\newblock Simultaneous detection and segmentation.
\newblock In \emph{Computer Vision--ECCV 2014: 13th European Conference, Zurich, Switzerland, September 6-12, 2014, Proceedings, Part VII 13}, 297--312. Springer.

\bibitem[{Hong et~al.(2022)Hong, Cho, Nam, Lin, and Kim}]{VAT}
Hong, S.; Cho, S.; Nam, J.; Lin, S.; and Kim, S. 2022.
\newblock Cost aggregation with 4d convolutional swin transformer for few-shot segmentation.
\newblock In \emph{European Conference on Computer Vision}, 108--126. Springer.

\bibitem[{Huang et~al.(2023)Huang, Cao, Li, Juefei-Xu, Lin, Tsang, Liu, and Guo}]{SAMrobust}
Huang, Y.; Cao, Y.; Li, T.; Juefei-Xu, F.; Lin, D.; Tsang, I.~W.; Liu, Y.; and Guo, Q. 2023.
\newblock On the robustness of segment anything.
\newblock \emph{arXiv preprint arXiv:2305.16220}.

\bibitem[{Kirillov et~al.(2023)Kirillov, Mintun, Ravi, Mao, Rolland, Gustafson, Xiao, Whitehead, Berg, Lo et~al.}]{SAM}
Kirillov, A.; Mintun, E.; Ravi, N.; Mao, H.; Rolland, C.; Gustafson, L.; Xiao, T.; Whitehead, S.; Berg, A.~C.; Lo, W.-Y.; et~al. 2023.
\newblock Segment anything.
\newblock In \emph{Proceedings of the IEEE/CVF International Conference on Computer Vision}, 4015--4026.

\bibitem[{Lang et~al.(2022)Lang, Cheng, Tu, and Han}]{BAM}
Lang, C.; Cheng, G.; Tu, B.; and Han, J. 2022.
\newblock Learning what not to segment: A new perspective on few-shot segmentation.
\newblock In \emph{Proceedings of the IEEE/CVF conference on computer vision and pattern recognition}, 8057--8067.

\bibitem[{Li et~al.(2023)Li, Zhang, Sun, Zou, Liu, Yang, Li, Zhang, and Gao}]{Semantic-SAM}
Li, F.; Zhang, H.; Sun, P.; Zou, X.; Liu, S.; Yang, J.; Li, C.; Zhang, L.; and Gao, J. 2023.
\newblock Semantic-sam: Segment and recognize anything at any granularity.
\newblock \emph{arXiv preprint arXiv:2307.04767}.

\bibitem[{Li et~al.(2021)Li, Jampani, Sevilla-Lara, Sun, Kim, and Kim}]{ASGNet}
Li, G.; Jampani, V.; Sevilla-Lara, L.; Sun, D.; Kim, J.; and Kim, J. 2021.
\newblock Adaptive prototype learning and allocation for few-shot segmentation.
\newblock In \emph{Proceedings of the IEEE/CVF conference on computer vision and pattern recognition}, 8334--8343.

\bibitem[{Lin et~al.(2014)Lin, Maire, Belongie, Hays, Perona, Ramanan, Doll{\'a}r, and Zitnick}]{coco}
Lin, T.-Y.; Maire, M.; Belongie, S.; Hays, J.; Perona, P.; Ramanan, D.; Doll{\'a}r, P.; and Zitnick, C.~L. 2014.
\newblock Microsoft coco: Common objects in context.
\newblock In \emph{Computer Vision--ECCV 2014: 13th European Conference, Zurich, Switzerland, September 6-12, 2014, Proceedings, Part V 13}, 740--755. Springer.

\bibitem[{Liu et~al.(2020)Liu, Zhang, Zhang, and He}]{ppnet}
Liu, Y.; Zhang, X.; Zhang, S.; and He, X. 2020.
\newblock Part-Aware Prototype Network for Few-Shot Semantic Segmentation.
\newblock In \emph{European Conference on Computer Vision}, 142--158.

\bibitem[{Ma et~al.(2024)Ma, He, Li, Han, You, and Wang}]{SAMmedical}
Ma, J.; He, Y.; Li, F.; Han, L.; You, C.; and Wang, B. 2024.
\newblock Segment anything in medical images.
\newblock \emph{Nature Communications}, 15(1): 654.

\bibitem[{Min, Kang, and Cho(2021)}]{HSNet}
Min, J.; Kang, D.; and Cho, M. 2021.
\newblock Hypercorrelation squeeze for few-shot segmentation.
\newblock In \emph{Proceedings of the IEEE/CVF international conference on computer vision}, 6941--6952.

\bibitem[{Nguyen and Todorovic(2019)}]{cocofewshot}
Nguyen, K.; and Todorovic, S. 2019.
\newblock Feature weighting and boosting for few-shot segmentation.
\newblock In \emph{Proceedings of the IEEE/CVF International Conference on Computer Vision}, 622--631.

\bibitem[{Park et~al.(2024)Park, Lee, Hyun, Seong, and Heo}]{TBS}
Park, S.; Lee, S.; Hyun, S.; Seong, H.~S.; and Heo, J.-P. 2024.
\newblock Task-Disruptive Background Suppression for Few-Shot Segmentation.
\newblock In \emph{Proceedings of the AAAI Conference on Artificial Intelligence}, volume~38, 4442--4449.

\bibitem[{Peng et~al.(2023)Peng, Tian, Wu, Wang, Liu, Su, and Jia}]{HDMNet}
Peng, B.; Tian, Z.; Wu, X.; Wang, C.; Liu, S.; Su, J.; and Jia, J. 2023.
\newblock Hierarchical dense correlation distillation for few-shot segmentation.
\newblock In \emph{Proceedings of the IEEE/CVF Conference on Computer Vision and Pattern Recognition}, 23641--23651.

\bibitem[{Radford et~al.(2021)Radford, Kim, Hallacy, Ramesh, Goh, Agarwal, Sastry, Askell, Mishkin, Clark et~al.}]{CLIP}
Radford, A.; Kim, J.~W.; Hallacy, C.; Ramesh, A.; Goh, G.; Agarwal, S.; Sastry, G.; Askell, A.; Mishkin, P.; Clark, J.; et~al. 2021.
\newblock Learning transferable visual models from natural language supervision.
\newblock In \emph{International conference on machine learning}, 8748--8763. PMLR.

\bibitem[{Rao et~al.(2022)Rao, Zhao, Chen, Tang, Zhu, Huang, Zhou, and Lu}]{DenseCLIP}
Rao, Y.; Zhao, W.; Chen, G.; Tang, Y.; Zhu, Z.; Huang, G.; Zhou, J.; and Lu, J. 2022.
\newblock Denseclip: Language-guided dense prediction with context-aware prompting.
\newblock In \emph{Proceedings of the IEEE/CVF conference on computer vision and pattern recognition}, 18082--18091.

\bibitem[{Roy et~al.(2023)Roy, Wald, Koehler, Rokuss, Disch, Holzschuh, Zimmerer, and Maier-Hein}]{SAMmd1}
Roy, S.; Wald, T.; Koehler, G.; Rokuss, M.~R.; Disch, N.; Holzschuh, J.; Zimmerer, D.; and Maier-Hein, K.~H. 2023.
\newblock Sam. md: Zero-shot medical image segmentation capabilities of the segment anything model.
\newblock \emph{arXiv preprint arXiv:2304.05396}.

\bibitem[{Shaban et~al.(2017{\natexlab{a}})Shaban, Bansal, Liu, Essa, and Boots}]{shaban2017one}
Shaban, A.; Bansal, S.; Liu, Z.; Essa, I.; and Boots, B. 2017{\natexlab{a}}.
\newblock One-shot learning for semantic segmentation.
\newblock \emph{arXiv preprint arXiv:1709.03410}.

\bibitem[{Shaban et~al.(2017{\natexlab{b}})Shaban, Bansal, Liu, Essa, and Boots}]{shaban2017one-shot_fss}
Shaban, A.; Bansal, S.; Liu, Z.; Essa, I.; and Boots, B. 2017{\natexlab{b}}.
\newblock One-shot learning for semantic segmentation.
\newblock \emph{arXiv preprint arXiv:1709.03410}.

\bibitem[{Shi et~al.(2022)Shi, Wei, Zhang, Lu, Ning, Chen, Ma, and Zheng}]{DCAMA}
Shi, X.; Wei, D.; Zhang, Y.; Lu, D.; Ning, M.; Chen, J.; Ma, K.; and Zheng, Y. 2022.
\newblock Dense cross-query-and-support attention weighted mask aggregation for few-shot segmentation.
\newblock In \emph{European Conference on Computer Vision}, 151--168. Springer.

\bibitem[{Strudel et~al.(2021)Strudel, Garcia, Laptev, and Schmid}]{segmenter}
Strudel, R.; Garcia, R.; Laptev, I.; and Schmid, C. 2021.
\newblock Segmenter: Transformer for semantic segmentation.
\newblock In \emph{Proceedings of the IEEE/CVF International Conference on Computer Vision}, 7262--7272.

\bibitem[{Sun et~al.(2024)Sun, Chen, Zhang, Zhang, Chen, Zhang, Ding, Wang, and Li}]{VRP-SAM}
Sun, Y.; Chen, J.; Zhang, S.; Zhang, X.; Chen, Q.; Zhang, G.; Ding, E.; Wang, J.; and Li, Z. 2024.
\newblock VRP-SAM: SAM with visual reference prompt.
\newblock In \emph{Proceedings of the IEEE/CVF Conference on Computer Vision and Pattern Recognition}, 23565--23574.

\bibitem[{Tang, Xiao, and Li(2023)}]{SAM-can}
Tang, L.; Xiao, H.; and Li, B. 2023.
\newblock Can sam segment anything? when sam meets camouflaged object detection.
\newblock \emph{arXiv preprint arXiv:2304.04709}.

\bibitem[{Tian et~al.(2020)Tian, Zhao, Shu, Yang, Li, and Jia}]{PFENet}
Tian, Z.; Zhao, H.; Shu, M.; Yang, Z.; Li, R.; and Jia, J. 2020.
\newblock Prior guided feature enrichment network for few-shot segmentation.
\newblock \emph{IEEE transactions on pattern analysis and machine intelligence}, 44(2): 1050--1065.

\bibitem[{Wang et~al.(2019)Wang, Liew, Zou, Zhou, and Feng}]{Panet}
Wang, K.; Liew, J.~H.; Zou, Y.; Zhou, D.; and Feng, J. 2019.
\newblock Panet: Few-shot image semantic segmentation with prototype alignment.
\newblock In \emph{proceedings of the IEEE/CVF international conference on computer vision}, 9197--9206.

\bibitem[{Wang, Sun, and Zhang(2023)}]{abcnet}
Wang, Y.; Sun, R.; and Zhang, T. 2023.
\newblock Rethinking the Correlation in Few-Shot Segmentation: A Buoys View.
\newblock In \emph{Proceedings of the IEEE/CVF Conference on Computer Vision and Pattern Recognition}, 7183--7192.

\bibitem[{Williams(2010)}]{pascal}
Williams, E. M. V. G.~L. 2010.
\newblock CK Winn J Zisserman A The pascal visual object classes (VOC) challenge.
\newblock \emph{Int. J. Comput. Vis}, 88(2): 303.

\bibitem[{Xie et~al.(2021)Xie, Wang, Yu, Anandkumar, Alvarez, and Luo}]{segformer}
Xie, E.; Wang, W.; Yu, Z.; Anandkumar, A.; Alvarez, J.~M.; and Luo, P. 2021.
\newblock SegFormer: Simple and efficient design for semantic segmentation with transformers.
\newblock \emph{Advances in Neural Information Processing Systems}, 34: 12077--12090.

\bibitem[{Xu et~al.(2025)Xu, Lin, Loy, Long, Li, and Zhao}]{ea}
Xu, Q.; Lin, G.; Loy, C.~C.; Long, C.; Li, Z.; and Zhao, R. 2025.
\newblock Eliminating feature ambiguity for few-shot segmentation.
\newblock In \emph{European Conference on Computer Vision}, 416--433. Springer.

\bibitem[{Xu et~al.(2024)Xu, Liu, Zhu, Lin, Long, Li, and Zhao}]{mambafewshot}
Xu, Q.; Liu, X.; Zhu, L.; Lin, G.; Long, C.; Li, Z.; and Zhao, R. 2024.
\newblock Hybrid mamba for few-shot segmentation.
\newblock \emph{arXiv preprint arXiv:2409.19613}.

\bibitem[{Xu et~al.(2023)Xu, Zhao, Lin, and Long}]{sccan}
Xu, Q.; Zhao, W.; Lin, G.; and Long, C. 2023.
\newblock Self-calibrated cross attention network for few-shot segmentation.
\newblock In \emph{Proceedings of the IEEE/CVF International Conference on Computer Vision}, 655--665.

\bibitem[{Yu et~al.(2023)Yu, Feng, Feng, Liu, Jin, Zeng, and Chen}]{InpaintAny}
Yu, T.; Feng, R.; Feng, R.; Liu, J.; Jin, X.; Zeng, W.; and Chen, Z. 2023.
\newblock Inpaint anything: Segment anything meets image inpainting.
\newblock \emph{arXiv preprint arXiv:2304.06790}.

\bibitem[{Zhang et~al.(2021)Zhang, Kang, Yang, and Wei}]{CyCTR}
Zhang, G.; Kang, G.; Yang, Y.; and Wei, Y. 2021.
\newblock Few-shot segmentation via cycle-consistent transformer.
\newblock \emph{Advances in Neural Information Processing Systems}, 34: 21984--21996.

\bibitem[{Zhang et~al.(2020)Zhang, Wei, Yang, and Huang}]{Sgone}
Zhang, X.; Wei, Y.; Yang, Y.; and Huang, T.~S. 2020.
\newblock Sg-one: Similarity guidance network for one-shot semantic segmentation.
\newblock \emph{IEEE transactions on cybernetics}, 50(9): 3855--3865.

\bibitem[{Zhou, Loy, and Dai(2022)}]{MaskCLIP}
Zhou, C.; Loy, C.~C.; and Dai, B. 2022.
\newblock Extract free dense labels from clip.
\newblock In \emph{European Conference on Computer Vision}, 696--712. Springer.

\end{thebibliography}

\newpage
\appendix
\onecolumn
\section{Appendix}
\subsection{Additional Qualitative Results}
\begin{figure*}[h!]
    {\includegraphics[width=1.0\textwidth]{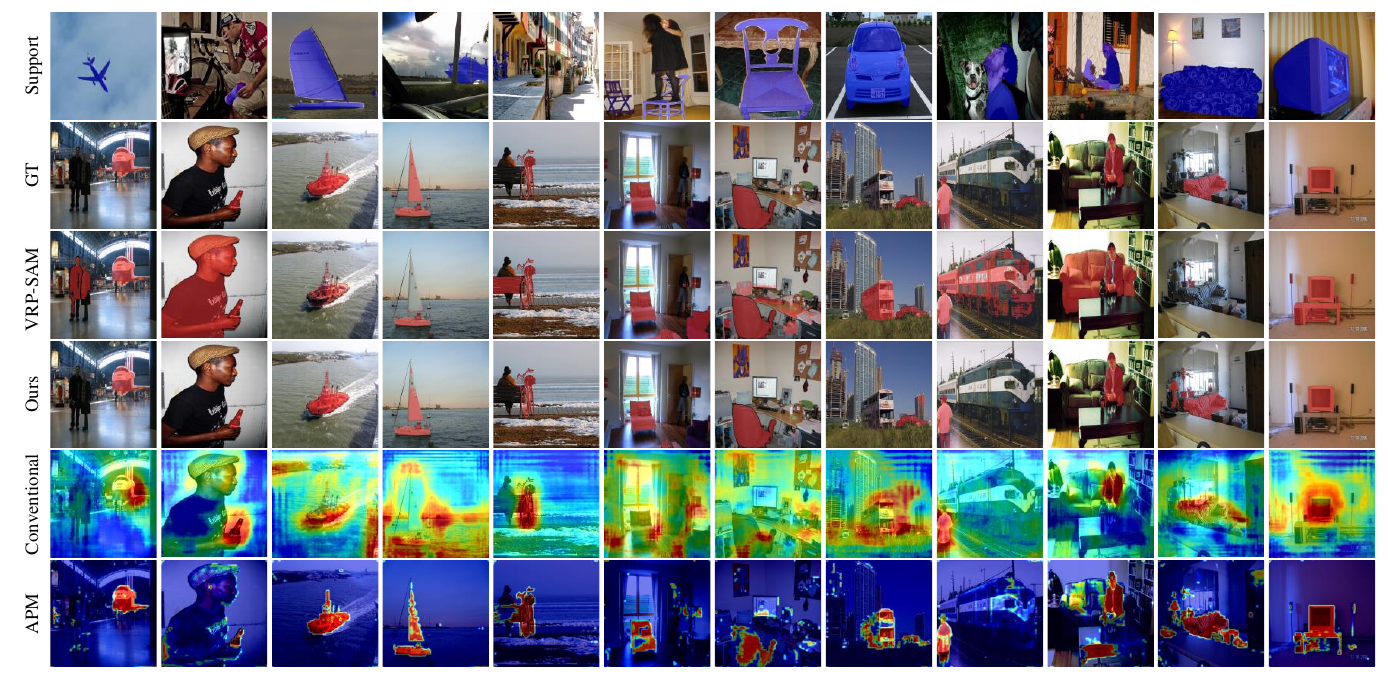}}
    \centering
    \vspace{-.75cm}
    \caption{
    Qualitative comparison results of Ours and VRP-SAM on the PASCAL-5$^i$ dataset. Conventional and APM refer to the conventional pseudo-mask of VRP-SAM and our Attention-based Pseudo-Mask, respectively.
    }
\label{fig:additional_qualitative}
\end{figure*}

Figure~\ref{fig:additional_qualitative} shows additional qualitative results and pseudo masks from baseline (VRP-SAM) and our method.
\vspace{-.25cm}

\vspace{-.5cm}
\subsection{Detailed Formulas}
\label{sec:appendix_formula}
We expand upon the formula briefly outlined in the manuscript.
First, we describe the computation process of $P^S_{t} \in \mathbb{R}^{C \times N}$ that the output of the masked cross-attention in Eq.~2.
This process utilizes $P^S_{t-1} \in \mathbb{R}^{C \times N}$, $\bar{G}^S \in \mathbb{R}^{C \times H \times W}$, and $M^S \in \mathbb{R}^{ H \times W}$, and is expressed as follows:
\vspace{-.2cm}
\begin{equation}
    \begin{split}
        \label{eq:masked_cross_attention} 
        \text{MaskedAttnWeight} \left( p_\textit{Q}(P^S_{t-1,n}), p_\textit{K}(\bar{G}^S_{h,w}); M^S_{h,w} \right) =& \frac {M^S_{h,w} \text{exp} \left( p_\textit{Q}(P^S_{t-1,n}) \odot p_\textit{K}(\bar{G}^S_{h,w}) / \sqrt{C}\right)} {\sum^H_{h'=1}\sum^W_{w'=1} M^S_{h',w'} \text{exp} \left( p_\textit{Q}(P^S_{t-1,n}) \odot p_\textit{K}(\bar{G}^S_{h',w'} ) / \sqrt{C}\right)},
        \\
        \vspace{-.1cm}
        P^S_{t,n} = \sum^H_{h=1}\sum^W_{w=1} \text{MaskedAttnWeight}& \left( p_\textit{Q}(P^S_{t-1,n}), p_\textit{K}(\bar{G}^S_{h,w}); M^S_{h,w} \right) \odot  p_\textit{V}(\bar{G}^S_{h,w}),
    \end{split}
\end{equation}
where $p_\textit{Q}(\cdot)$, $p_\textit{K}(\cdot)$, and $p_\textit{V}(\cdot)$ are projection layers for {\it query}, {\it key}, and {\it value}.
Next, we detail the computation process for the cross-attention output $P^Q_t \in \mathbb{R}^{C \times N}$ from Eq.~7.
It can be formulated with $P^Q_{t-1} \in \mathbb{R}^{C \times N}$ and $\bar{G}^Q \in \mathbb{R}^{C \times H \times W}$, as follows:
\begin{equation}
    \begin{split}
        \label{eq:cross_attention} 
        \text{AttnWeight} \left( p_\textit{Q}(P^Q_{t-1,n}), p_\textit{K}(\bar{G}^Q_{h,w}) \right) =& \frac {\text{exp} \left( p_\textit{Q}(P^Q_{t-1,n}) \odot p_\textit{K}(\bar{G}^Q_{h,w} ) / \sqrt{C}\right)} {\sum^H_{h'=1}\sum^W_{w'=1} \text{exp} \left( p_\textit{Q}(P^Q_{t-1,n}) \odot p_\textit{K}(\bar{G}^Q_{h',w'} ) / \sqrt{C}\right)},
        \\
        \vspace{-.1cm}
        P^Q_{t,n} = \sum^H_{h=1}\sum^W_{w=1} \text{AttnWeight}& \left( p_\textit{Q}(P^Q_{t-1,n}), p_\textit{K}(\bar{G}^Q_{h,w})\right) \odot  p_\textit{V}(\bar{G}^Q_{h,w}).
    \end{split}
\end{equation}

Further, we describe the binary cross-entropy and dice loss, utilized in Eq.~9 and 13.
Suppose that we are given a pseudo mask $M^{\text{pred}} \in \mathbb{R}^{H \times W}$ and the corresponding ground-truth mask $M^{\text{gt}} \in \mathbb{R}^{H \times W}$, the binary cross-entropy loss between them $\mathcal{L}_{\text{BCD}}(M^\text{pred}, M^{\text{gt}})$ is defined as follows:
\vspace{-.1cm}
\begin{equation}
    \begin{split}
        \label{eq:loss_BCE} 
        \mathcal{L}_{\text{BCD}}(M^\text{pred}, M^\text{gt}) = - \frac{1}{HW} \sum^{H}_h \sum^{W}_w \left( M^\text{gt}_{h,w} \text{log}\left(M^\text{pred}_{h,w} \right) + (1 - M^\text{gt}_{h,w}) \text{log}\left(1 - M^\text{pred}_{h,w} \right) \right)
    \end{split}
\end{equation}
Meanwhile, the dice loss between them $\mathcal{L}_{\text{DL}}(M^\text{pred}, M^{\text{gt}})$ is formulated as follows:
\begin{equation}
    \begin{split}
        \label{eq:loss_DL} 
        \mathcal{L}_{\text{DL}}(M^\text{pred}, M^\text{gt}) = 1 - \frac{2\sum^{H}_{h=1} \sum^{W}_{w=1} M^\text{gt}_{h,w} M^\text{pred}_{h,w}} 
        {\sum^{H}_{h=1} \sum^{W}_{w=1} M^\text{gt}_{h,w} M^\text{gt}_{h,w} + \sum^{H}_{h=1} \sum^{W}_{w=1} M^\text{pred}_{h,w} M^\text{pred}_{h,w}}
        .
    \end{split}
\end{equation}

\end{document}